\documentclass[final]{cvpr}

\usepackage{times}
\usepackage{epsfig}
\usepackage{graphicx}
\usepackage{amsmath}
\usepackage{amssymb}

\usepackage{multicol}
\usepackage{multirow}
\usepackage{graphics}
\usepackage{booktabs}
\usepackage{caption}
\usepackage{subcaption}
\usepackage[pagebackref=true,breaklinks=true,colorlinks,bookmarks=false]{hyperref}

\def\cvprPaperID{3978} 
\def\confYear{CVPR 2021}

\begin{document}
\pagestyle{empty}  


\title{Complementary Relation Contrastive Distillation}
\renewcommand{\thefootnote}{\fnsymbol{footnote}}
\author{Jinguo Zhu$^1$\footnotemark[2] \qquad Shixiang Tang$^2$ \qquad Dapeng Chen$^3$\footnotemark[3]  \qquad Shijie Yu$^4$ \\
Yakun Liu$^3$   \qquad Mingzhe Rong$^1$  \qquad Aijun Yang$^1$   \qquad Xiaohua Wang$^{1}$\footnotemark[3] \\
$^1$Xi'an Jiaotong University  \qquad $^2$The University of Sydney \qquad  $^3$Sensetime Group Limited \\  
$^4$Shenzhen Institutes of Advanced Technology, CAS\\
{\tt\small lechatelia@stu.xjtu.edu.cn \quad  tangshixiang@sensetime.com \quad  384822707@qq.com }
\\ 
{\tt\small
 sj.Yu@siat.ac.cn \quad liuyakun1@sensetime.com \quad \{mzrong, yangaijun, xhw\}@mail.xjtu.edu.cn
}
}

\maketitle
\thispagestyle{empty} 
\footnotetext[2]{This work was done when Jinguo Zhu was an intern at SenseTime.}
\footnotetext[3]{Corresponding authors.}

\begin{abstract}
Knowledge distillation aims to transfer representation ability from a teacher model to a student model.  Previous approaches focus on either individual representation distillation or inter-sample similarity preservation. While  we argue that the inter-sample relation conveys abundant information  and needs to be distilled in a more effective way.  
In this paper, we propose a novel knowledge distillation method, namely Complementary Relation Contrastive Distillation (CRCD), to transfer the structural knowledge from the teacher to the student. Specifically,  we estimate the mutual relation in an anchor-based way and distill the anchor-student relation under the supervision of its corresponding anchor-teacher relation. To make it more robust, mutual relations are modeled by two complementary elements: the feature and its gradient. Furthermore, the low bound of mutual information between the anchor-teacher relation distribution  and the anchor-student relation distribution is maximized via relation contrastive loss, which can distill both the sample representation and the inter-sample relations. Experiments on different benchmarks demonstrate the effectiveness of our proposed CRCD.



\end{abstract}

\section{Introduction}
\begin{figure}[t]
\begin{center}
    \includegraphics[width=240pt]{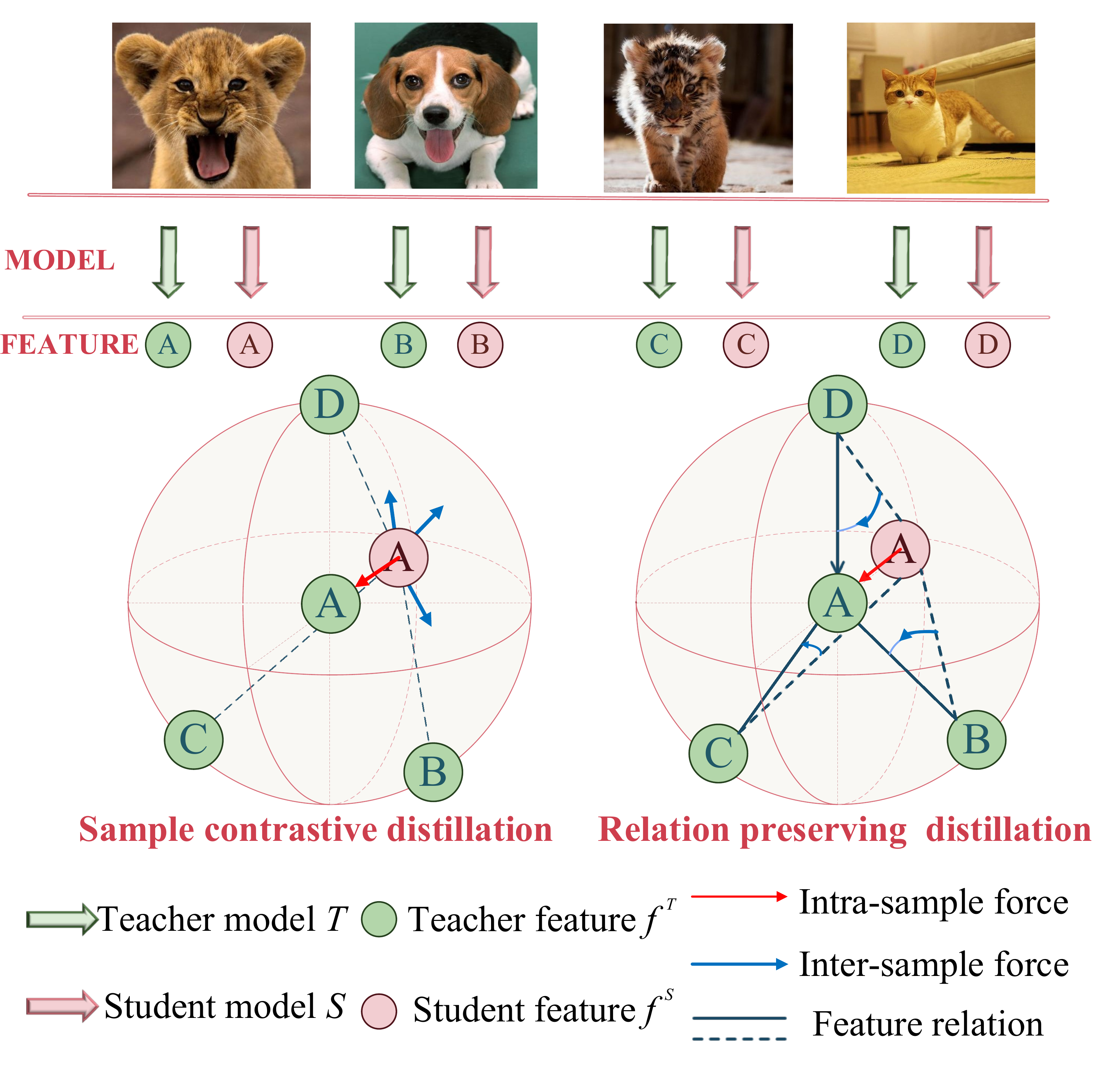}
    \caption{ Sample contrastive distillation \textit{vs.} Relation preserving distillation. 
    Four neighboring samples and their corresponding features are displayed, and 
    capital letters are used to identify them.
    While pulling  $f_A^S$ closer to $f_A^T$,  sample contrastive distillation  will simultaneously push $f_A^S$ away from $f_B^T$, $f_C^T$ and $f_D^T$ without distinction,
    whereas relation preserving distillation preserves the feature relations across the feature space, thus $f_A^S$ can be optimized along the optimal direction.
    }
    \label{fig:challenge}
\end{center}
\vspace{-1.8em}
\end{figure}

Knowledge distillation aims to transfer the knowledge from one deep learning model (the teacher) to another (the student), such as distilling a large network into a smaller one \cite{hinton2015distilling,yim2017gift,belagiannis2018adversarial,xu2017training,ge2020mutual} or ensembling a collection of models into a single model \cite{malinin2019ensemble,shen2019meal,liu2019knowledge,wu2019distilled}.  It has a wide range of applications in the industry especially when a neural network needs to be efficiently deployed on devices with limited computational resources \cite{chen2020big,crowded,MutualCRF}.  Although great progress has been achieved in the knowledge distillation regime, there is still no consensus on what kind of knowledge really needs to be preserved in the distillation \cite{gou2020knowledge}.

As one of the most effective distillation methods, CRD \cite{crd} holds the view that the representational knowledge is structured. So It tries to capture the correlations and higher-order output dependencies for each sample, which is different from the original KD objective introduced in \cite{hinton2015distilling} that treats all dimensions as independent information. CRD leverages the family of contrastive objectives \cite{DeepInfoMAx,Contrastive_multiview_coding, wu2018unsupervised, caron2020unsupervised} to maximize a lower-bound of the mutual information between the teacher and student representations. It essentially performs knowledge distillation based on the individual samples, enforcing the representation consistency between the teacher model and the student model. 

However, neither CRD nor other sample-based distillation methods can effectively preserve inter-sample relations, which are more valuable than the sample representations themselves in many practical tasks, \emph{e.g.,} retrieval and classification.  As shown in Fig.  \ref{fig:challenge}, 
%
when using sample contrastive distillation methods, \emph{e.g.,} CRD,
the optimized forces from other neighbors just push the student representation of sample $A$ away when contrasted negatively,
which may not be optimal and can break the latent structural geometry of neighboring samples. 
Some recent works have shown that transferring the mutual similarity instead of actual representation is beneficial to student representation learning \cite{tung2019similarity,park2019relational,peng2019correlation,pkt}. 
These methods directly estimate the relations in teacher space by computing the inter-sample similarities, then mimic these similarities in the student space via $L_2$ loss or $KL$ divergence, ignoring the high-order dependency within the representation in both relation estimation and knowledge distillation.  

To robustly distill the  structural knowledge of the teacher space, we define a new cross-space relation between two samples and supervise this new relation by its corresponding relation in the teacher representation space. More specifically, given the teacher and student representation of one sample, we select  a neighboring sample's representation  from the teacher representation space as an anchor. The anchor-student relation is encouraged to be consistent with the anchor-teacher relation. Our method brings at least three merits for distillation.  
(1) It simultaneously optimizes the representation and relation. When the anchor-student relation is pushed to be consistent with the anchor-teacher relation, the student representation is actually optimized along the optimal direction of representation learning.
(2) The anchor-student relation is more effective for distillation compared with the student-student relation (where two representations are both from the student space) in the conventional KD family \cite{tung2019similarity,park2019relational,peng2019correlation}. The student-student relation is unstable because the two representations in the student space are not well optimized and they will drift significantly during distillation, while the anchor representation within the anchor-student relation is fixed, which can effectively optimize the representation in the student space. 
(3) As the anchor can be randomly selected from the neighborhood of the considered sample, the student representation of one sample is supervised by multiple relations from different anchors, which guarantees the robustness of the distillation. 



The representation relation is modeled by two complementary elements: the feature and its gradient. The feature relation reflects the structural information in representational space, and the gradient relation is computed by the feature gradients after backward propagation. As gradients measure the fastest rate and direction for loss minimization, gradient relation can explore the structural information of optimization kinetics in representational space \cite{ab,LayerwiseCTL}. During the distillation, we maximize the mutual information between the anchor-teacher relation and the anchor-student relation for both two elements. The maximization problem can further surrogate to maximize the lower bound of mutual information which has been well solved by contrastive learning \cite{wu2018unsupervised}. Our method is therefore denoted by Complementary Relation Contrastive Distillation(CRCD). 

In summary, the main contributions of CRCD are three-fold. First, we define a new  anchor-based cross space relation and adopt it to effectively and robustly distill both sample representations and inter-sample relations. Second, the new relation is modeled by two complementary elements, \emph{i.e.}, the feature and its gradients, which capture the structure information of the feature and the optimization kinetics, respectively. Last, we maximize the low bound of mutual information between the anchor-teacher relation and the anchor-student relation and derive an efficient solution in the form of contrastive learning.  Extensive experiments empirically validate the effectiveness of CRCD and further improve the current state-of-the-art in various benchmarks.

\section{Related Work}
\begin{figure*}[t]
\begin{center}
    \includegraphics[width=480pt]{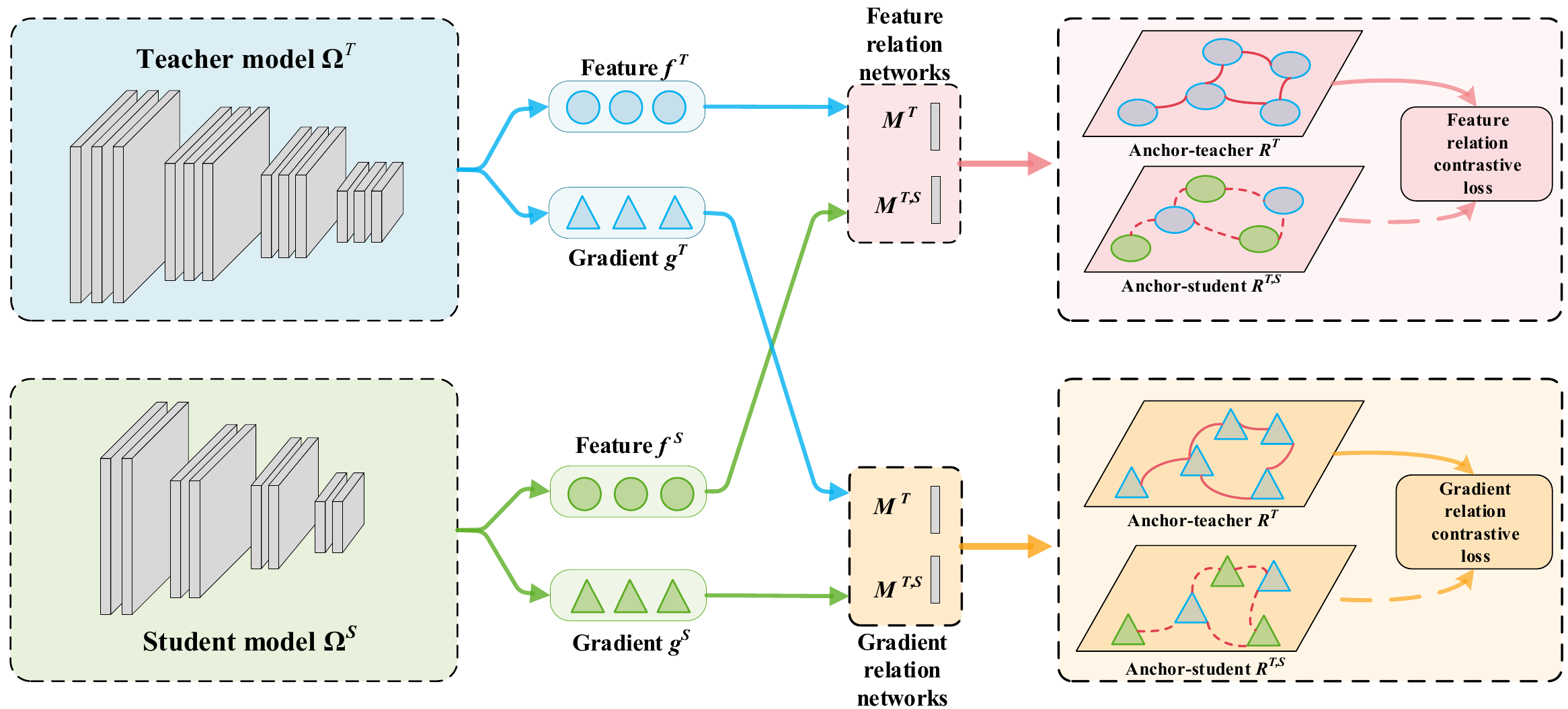}
    \caption{ \textbf{The Flowchart of CRCD.}
    To distill the structural knowledge from the teacher model $\Omega^T$ to the student model $\Omega^S$, 
     two complementary elements,
    the feature $f$ and its gradient $g$, are utilized to model the representation relations.
    For each element, auxiliary subnetworks $M^T$ and $M^{T,S}$ are used to estimate the anchor-teacher relation $R^{T}$ in the teacher space and anchor-student relation $R^{T,S}$ across space respectively.
    Meanwhile, the cross-space $R^{T,S}$ is supervised by its corresponding $R^{T}$. 
    By this way, not only the relation estimation but also the representation learning can be achieved.
    }
    \label{fig:flowchart}
    
\end{center}
\vspace{-1.8em}
\end{figure*}

\noindent \textbf{Knowledge Distillation.}
There has been a rising interest in distilling knowledge from one model to another, in which the core issue is that what is the knowledge learned by a teacher and how to best distill the knowledge into a student. 
In \cite{hinton2015distilling}, the soft probability distribution is transferred by using a higher temperature value.
Compared to the one-hot label, soft targets can contain much more valuable information that defines a rich similarity structure over the data.
Furthermore, not only the soft labels but also the hints from intermediate layers are used to train student networks in \cite{romero2014fitnets}.
Moreover, the attention map \cite{ATloss} and the flow of solution procedure (FSP) \cite{fsp} are used to transfer knowledge between networks.
These works focus on distilling the knowledge modeled by learned presentations of samples themselves, 
however, ignore the mutual relations between samples, which contain rich structural information learned by the teacher.

There are a few recent works analyzing and exploiting the mutual relation between data samples \cite{relationgraph,peng2019correlation,park2019relational,pkt,Chen_2018_CVPR,Chen_2018_CVPR_video,Chen_2016_CVPR,Li_2019_ICCV}.
In particular, similarity-preserving knowledge   \cite{tung2019similarity} proposes to transfer the knowledge presented as similar activation between input pairs.
In \cite{peng2019correlation} and \cite{park2019relational}, the sample relations are modeled explicitly to transfer knowledge.
However,  these methods all use low-dimensional relation methods, such as cosine similarity \cite{tung2019similarity}  or gaussian RBF \cite{peng2019correlation} between features, to model the mutual relation, which may be suboptimal for modeling complex inter-sample interdependencies.
Instead, in our paper, we  design  sub-networks to learn the  high-dimensional across-space relations which can capture the complex mutual dependencies of deep representations from any two feature spaces.

\noindent \textbf{Contrastive Learning.}
Contrastive Learning serves as the core idea of several recent works on self-supervised representation learning \cite{SimCLR, moco, CPC, DeepInfoMAx,byol,wang2020understanding,tian2020makes,ge2020selfpaced}.
Contrastive losses such as NCE \cite{infoNCE,DeepInfoMAx} measure the similarities of data samples in a deep representation space, which learn representation by contrasting positive and negative representation pairs.
For knowledge distillation, CRD \cite{crd} is the first study that combines contrastive learning with  knowledge distillation, which aims to maximize mutual information \cite{mine} between the teacher and student representations.
Besides, SSKD \cite{SSKD} proposes to use contrastive tasks as self-supervised pretext tasks, which can facilitate the extraction of richer knowledge from the teacher to the student.
From the usage of the contrastive loss, our method is more similar to CRD, but our objective is the mutual relations of deep representations, instead of the representations themselves. 

\section{Methodology}

Fig. 1 presents the overall flowchart of our proposed CRCD.
Given a teacher network $\Omega^{T}$ and a student network $\Omega^{S}$, we denote the representation of an input $x$ produced by the two networks as $\phi^{T}(x)$ and $\phi^{S}(x)$, respectively.  
Let $x_{i}$ and $x_{j}$ be two training samples randomly chosen from  the sample set $X$.
We denote the relation in the teacher space as $r^{T}_{i,j}$, where $r^{T}_{i,j} $ is a vector computed by a sub-network $M^{T}$ that takes $\phi^{T}(x_{i})$ and $\phi^{T}(x_{j})$ as inputs. 
We further define a new relation $r^{T,S}_{i,j} $ computed by another sub-network $M^{T,S}$. It is noteworthy that the inputs  $\phi^{T}(x_{i})$ and $\phi^{S}(x_{j})$ for $M^{T,S}$ are from different spaces. 
Regarding 
$\phi^{T}(x_{i})$ 
as an anchor representation, the cross-space  anchor-student relation $r^{T,S}_{i,j}$ is expected to be consistent with the teacher-space anchor-teacher relation $r^{T}_{i,j}$, which not only preserves the relation between $x_{i}$ and $x_{j}$, but also drives the $\phi^{S}(x_{j})$ to be consistent with $\phi^{T}(x_{j})$. 

In the following sub-sections, we first demonstrate how to use contrastive learning to perform the relation distillation, then  two complementary elements are introduced to model the representation relations, and the implementation details and some discussions will be presented at last. The complete mathematical derivation refers to the supplementary materials.

\subsection{Relation Contrastive Distillation}
Assume that we are given a set of training examples with empirical data distribution $p(X)$, the sampling procedure for the conditional marginal distributions $p(R^{T}|X)$, $p(R^{T,S}|X)$ are modeled as
\begin{equation}
    \small 
     \begin{split}
      & x_{i} ,  x_{j} \sim p(X), \quad \ \   r^{T}_{i,j} = M^{T}(\phi^T(x_{i}), \phi^T(x_{j})),    \\
      & x_{m}, x_{n} \sim p(X), \quad r^{T,S}_{m,n} = M^{T, S}(\phi^T(x_{m}), \phi^S(x_{n}))\\
    \end{split} \label{eq:marginal}
\end{equation}
respectively.
While the sampling procedure of the conditional joint distribution $p(R^{T},R^{T,S}|X)$ is modeled as:
\begin{equation}
\small
\begin{split}
  x_{i}, x_{j} \sim p(X), &  \quad r^{T}_{i,j} = M^{T}(\phi^T(x_{i}), \phi^T(x_{j})), \\
 & \quad r^{T,S}_{i,j} = M^{T, S}(\phi^T(x_{i}), \phi^S(x_{j})).\\
\end{split} \label{eq:joint}
\end{equation}
For ease of notation, we utilize $p(R^{T})$,  $p(R^{T,S})$ and $p(R^{T}, R^{T,S})$ to briefly represent $p(R^{T}|X)$,  $p(R^{T,S}|X)$ and $p(R^{T},R^{T,S}|X)$. Intuitively, we aim to maximize the mutual information (MI) of the two relation distributions from $R^{T}$ and $R^{T,S}$,  which is


 \begin{equation}
     I(R^{T}, R^{T, S}) = \mathbb{E}_{p(R^{T},R^{T,S})} \log \frac{ p(R^{T},R^{T,S})}{p(R^{T})p(R^{T,S})}.  
 \end{equation}
 \noindent \textbf{MI Lower Bound.} \quad To derive a solvable loss function, we define a distribution $q$ with latent variable $C$ which indicates whether the relation tuple $(r^{T}, r^{T, S})$ is drawn from the joint distribution or the product of marginal distributions:
 \begin{equation} \small
   \begin{split}
     q(R^{T}, R^{T, S}| C=1) & = p(R^{T},R^{T,S}) \\
     q(R^{T}, R^{T, S}| C=0) & = p(R^{T})p(R^{T,S}).
\end{split}
 \end{equation}
More specifically, $C\!=\!1$ means $r^{T}$ and $r^{T, S}$ are computed based on the same input pair as in Eq.  \ref{eq:joint}, and $C\!=\!0$ means $r^{T}$ and  $r^{T, S}$ are independently selected as in Eq. \ref{eq:marginal}. In our data, we provide $1$ relevant relation pair ($C\!=\!1$) with $N$ irrelevant relation pair ($C\!=\!0$). Then the prior $q(C=1)= 1/(N+1)$ and $q(C=0)=N/(N+1)$.  Combing the priors with the Bayes' rule, the posterior for $C=1$ is given by:
\begin{equation}\small
    q(C\!=\!1|R^{T}, R^{T, S}) = \frac{p(R^{T}, R^{T, S})}{p(R^{T}, R^{T, S})+Np(R^{T})p(R^{T, S})}.
\end{equation}
By connection to the mutual information, the posterior $\log q(C\!=\!1|R^{T}, R^{T, S}) \leq	-\log(N)+\log\big(\frac{p(R^{T}, R^{T, S})}{p(R^{T})p(R^{T, S}) } \big)$.  Taking the expectation on both sides w.r.t. $p(R^{T}, R^{T, S})$, which is also equivalent to $ q(R^{T}, R^{T, S}| C=1)$, we have:
\begin{equation} \small
\begin{split}
     I(R^{T}, R^{T, S}) & \geq \log(N) + \\ 
                       & \mathbb{E}_{q(R^{T}, R^{T, S}| C=1)}\log q(C\!=\!1|R^{T}, R^{T, S})
\end{split}\label{Eq:mutual}
\end{equation}
where $\log(N) + \mathbb{E}_{q(R^{T}, R^{T, S}| C=1)}\log q(C\!=\!1|R^{T}, R^{T, S})$ is a lower bound of the mutual information.

 \noindent \textbf{Distribution Approximation.} \quad As there is no knowledge about the true distribution of $q(C\!=\!1|R^{T}, R^{T, S})$, we approximate 
 the distribution
 by fitting a parameterized model $h$: $\{R^{T},  R^{T, S}\} \rightarrow	[0, 1]$ with the samples from
$q(C\!=\!1|R^{T}, R^{T, S})$.
The log-likelihood of the sampled data under this model is defined as:
 \begin{equation} \small
 \begin{split}
\mathcal{I}(h) & = \mathbb{E}_{q(R^{T}, R^{T, S}| C=1)}[\log h(R^{T}, R^{T, S})] \\
                & + N\mathbb{E}_{q(R^{T}, R^{T, S}| C=0)}[\log(1 - h(R^{T}, R^{T, S}))].
\end{split} \label{critic}
 \end{equation}
To achieve a good approximation to $q(C=1|R^{T}, R^{T, S})$, we need to maximize the log likelihood.  Consider the bound in Eq. \ref{Eq:mutual} and the fact that
 $N\mathbb{E}_{q(R^{T}, R^{T, S}| C=0)}[\log(1 - h(R^{T}, R^{T, S}))]$ is non-positive, we have 
 \begin{equation} \small
     \begin{aligned}
        I(R^{T}, R^{T, S}) &   \geq   \log N \!+\! 
         \mathbb{E}_{q(R^{T}, R^{T, S}| C=1)}[\log h(R^{T}, R^{T, S})]  \\
        & + N\mathbb{E}_{q(R^{T}, R^{T, S}| C=0)}[\log(1 - h(R^{T}, R^{T, S}))] \\
        & \geq \log N + \mathcal{I}(h),
     \end{aligned} 
     \label{MI_eq}
 \end{equation}
where $\log N + \mathcal{I}(h)$ is the lower bound of the mutual information with the parameterized model $h$.  The maximization of the log-likelihood is also to maximize the lower bound.

\noindent \textbf{Relation Contrastive Loss.} \quad In our method, the inputs for the function  $h$ are teacher-space relation $r^T$ and cross-space relations $r^{T,S}$, which are the results of the teacher $\Omega^{T}$, the student $\Omega^{S}$, and the two sub-networks $M^{T}, M^{T,S}$. Except the teacher $\Omega^{T}$, the other three networks $ \Omega^{S}$, $M^{T}$ and $M^{T,S}$ also need to be optimized during the distillation. We aim to maximize the mutual information, which is equivalent to minimizing the relation contrastive loss $\mathcal{L}_{RC}$:
\begin{equation}\small
\begin{split}
    \mathcal{L}_{RC}(h, \Omega^{S}, M^{T}, M^{T,S})   & = 
    \!-\! \sum_{q(C=1)} \log h(r^{T}, r^{T, S}) \\
    & - N\sum_{q(C=0)}  \log[1 \!-\! h(r^{T}, r^{T, S})]
\end{split}
\label{contrastive_loss}
\end{equation}
where $\{(r^{T},  r^{T, S})|C=1\}$ act as positive pairs while $\{(r^{T},  r^{T, S})|C=0\}$ act as negative pairs. Due to Eq. \ref{MI_eq}, the contrastive loss can fit the distribution $q(C|R^{T}, R^{T, S})$ to increase the lower-bound of mutual information of $R^{T}$ and $R^{T, S}$, by which not only the parameterized model $h$, but also the other three networks $ \Omega^{S}$, $M^{T}$ and $M^{T,S}$ can be jointly optimized.


\subsection{Complementary Relation}
 Modeling relation between sample representations is the prerequisite for distilling the structural information. We therefore propose two learnable sub-networks $M^{T,S}$ and $M^{T}$ to estimate the relation. 

The sub-network $M^{T,S}$ is to compute the anchor-student relation with representation  $\phi^{T}(x_{i})$ and $\phi^{S}(x_{j})$:
\begin{equation} \small
    \begin{split}
        r^{T,S}_{i,j} & = M^{T,S}(\phi^{T}(x_{i}),\phi^{S}(x_{j})) \\
        & = W^{A}(\sigma(W_{i}^{A}\phi^{T}(x_{i}) -W_{j}^{A}\phi^{S}(x_{j}))), \\
    \end{split}
\label{action_cross_relation}
\end{equation}
where $W_i^{A}$ and $W_j^{A}$ are linear transformations that can solve the dimension mismatch problem. $\sigma$ is ReLU function and $W^{A}$ is used for transformation. The anchor-student relation is supervised by the fixed anchor-teacher relation $ r^{T}(x_i,x_j)$, computed by another sub-network $M^{T}$:
\begin{equation} \small
    \begin{split}
        r^{T}_{i,j} & = M^{T}(\phi^{T}(x_{i}),\phi^{T}(x_{j})) \\
        & = W^{B}(\sigma(W_i^{B}\phi^{T}(x_{i})-W_j^{B}\phi^{T}(x_{j}))). \\
    \end{split}
\label{action_relation}
\end{equation}
It is noteworthy that the relations $r^{T,S}$ and  $r^{T}$ are not scalar values but high-dimensional vectors. We claim that the high-dimensional relation can more accurately capture the structural information of deep representations than low-dimensional relation \emph{e.g.,} cosine similarity,
which will be validated in section \ref{exp_relation}.
Furthermore, the small learnable networks also increases relation flexibility.
%

The relations are modeled by two complementary elements: \textit{feature} $f$ and its \textit{gradient} $g$.
Specifically, the representation $\phi(x)$ in Eq.  \ref{action_cross_relation} and Eq.  \ref{action_relation} can be either the feature of the teacher/student model or its gradient.

\noindent \textbf{Feature Element.}
The feature element is the $\ell_2$ normalized output of teacher/student's backbone. With the feature element $f$, the representations $\phi^{T}(x)$ and $\phi^{S}(x)$ reflect the direct activation relative to the input $x$:
\begin{equation} \small
    \begin{split}
       \phi^{T}(x) = f^T(x) ;\quad   \phi^{S}(x) = f^S(x)
    \end{split}
\end{equation}

\noindent \textbf{Gradient Element.}
The gradient element is the gradient with respect to the feature. It reflects the optimization kinetics in the feature space, encoding important structural information. Given an input sample $x$ into a teacher/student network $\Omega$, the gradient of task loss $\mathcal{L}_{cls}$ relative to the feature $f$ is computed as:
\begin{equation} \small
     \mathit{g(x)} = \frac{\partial}{\partial f}\mathcal{L}_{cls}(\Omega, x).
\end{equation}
With gradient elements, the representation $\phi^{T}(x)$ and $\phi^{S}(x)$ can reflect the optimization kinetics:
\begin{equation} \small
    \begin{split}
       \phi^{T}(x) = g^T(x) ;\quad   \phi^{S}(x) = g^S(x)
    \end{split}
\end{equation}

\noindent \textbf{Element Combination.}
\textit{Complementary relation} is modeled to  leverage feature and gradient elements simultaneously.
Specifically,
after  the one-sided relations: \textit{feature relation} $r^f$ and \textit{gradient relation} $r^g$ , are computed with   feature and gradient elements respectively, their corresponding relation contrastive losses can also be calculated by Eq. \ref{contrastive_loss}.
By optimizing these two losses simultaneously, these two elements can both be utilized.


\subsection{Implementation}
\noindent \textbf{Critic Function.} We specify the parameterized critic function $h$ in Eq. \ref{critic} to  distinguish whether the relation pair $(r^{T}, r^{T,S})$ is sampled from the joint distribution $p(R^{T},R^{T,S})$ or the product of marginal distribution $p(R^{T})p(R^{T,S})$.  The formulation is similar to NCE \cite{wu2018unsupervised}:
\begin{equation}
\small
    \begin{split}
        h(r^{T},  r^{T, S}) = \frac{e^{h_1(r^{T})h_2(r^{T, S})/\tau}}{e^{1/\tau}}
    \end{split}
    \label{hfunction}
\end{equation}
where $\tau$ is a temperature hyperparameter, and $h_1$ and $h_2$ first perform the linear transformation on relations, then normalize the transformed relations with $\ell_2$ norm.


\noindent \textbf{Sampling Policy.} We adopt the following sampling policy:  in each forward-propagation, the anchor relation $r^T_{ij}$ and positive relation $r^{T,S}_{ij}$ are calculated using representations from any two samples $x_i$ and $x_j$ 
in the current mini-batch, while the negative relations $r^{T,S}_{ik}$ are calculated using the anchor representation from $x_{i}$ and the representations (indexed with $k$) sampled from the buffer   where features and gradients are stored. 
Considering a $B$-size min-batch, we construct the anchor/positive relation for each sample pair 
thus the number of these two relations can be $B^{2}$. For each anchor relation, we sample $N$ feature/gradient from the buffer to construct $N$ negatives for contrastive learning.  

To make the feature/gradient buffer reflect the current network state better, we propose a queuing sampling method instead of a randomly sampling strategy. The queue records the $N$ sample indices from the immediate preceding mini-batches and is updated  after each forward-propagation by replacing the oldest indices with the current mini-batch. According to these recorded indices, the representations of these samples are
used to calculated relation contrastive loss, 
whose effectiveness will be
studied in Sec. \ref{ablation}.

\noindent \textbf{Loss Function.}  To achieve the superior performance and conduct a fair comparison with other methods, we also incorporate the naive knowledge distillation loss $\mathcal{L}_{kd}$ \cite{hinton2015distilling} along with our relation contrastive loss. 
Given the pre-softmax logits $z^T$ and $z^S$ for teacher and student, the naive KD loss can be expressed as
\begin{equation}
    \small
    \mathcal{L}_{kd} = \rho^2  \mathcal{H}(\sigma(z^T/\rho),\sigma(z^S/\rho))
\end{equation}
where $\rho$ is the temperature, $\mathcal{H}$ refers to the cross-entropy and $\sigma$ is  softmax function. 
The complete objective is:
\begin{equation}
    \small
    \mathcal{L} =  \mathcal{L}_{cls} + \alpha \mathcal{L}_{KD} + \beta_1 \mathcal{L}_{RC}^{f} + \beta_2 \mathcal{L}_{RC}^{g}
\end{equation}
where $\mathcal{L}_{RC}^{f}$ and $\mathcal{L}_{RC}^{g}$ are the relation contrastive loss computed with the feature ($f$) and gradient ($g$), respectively. $\mathcal{L}_{cls}$ is the cross entropy loss for classification. We set hyper-parameters to $\alpha=1$ and $\beta_{1}=\beta_{2}=0.5$ empirically.  

\noindent \textbf{Discussion.}  CRD \cite{crd} aims to maximize the mutual information between the representations of the sample themselves from teacher/student models. Meanwhile, the proposed CRCD seeks the consistency between the teacher-space relation and cross-space relation. Indeed, if $i=j$ in Eq. \ref{contrastive_loss}, the loss of CRCD essentially optimizes the cross-space relation of one sample, which degrades to the loss of CRD. Moreover, the number of pair-wise relations is at quadratic level relative to the number of samples, which also increases the optimized stability of  contrastive  loss. 

\section{Experiments}

\subsection{Datasets and Experimental Setup}

\noindent \textbf{Datasets.} Our experiments are conducted on two widely used classification datasets, \emph{i.e.}, CIFAR100 \cite{cifar100} and ImageNet \cite{imagenet}. CIFAR100 contains 60000 images for 100 classes, and there are 500 and 100 images per class for training and testing respectively. ImageNet is a well-known large-scale image classification benchmark with 1000 classes, consisting of 1281167 images for training and 50000 images for testing.


\begin{table}[t]
\small
    \centering
     \caption{Testing accuracy (\%) on CIFAR100 with different relation modeling methods. 
     $\mathcal{L}_2$ loss and relation contrastive loss $\mathcal{L}_{RC}$ are used to distill the feature relation $r^f$.
    } 
\begin{tabular}{c|cc|c}
\toprule
teacher   & resnet56 &resnet110 &  ResNet50 \\
student & resnet20 & resnet20 & vgg8\\

\midrule

 RKD \cite{park2019relational} & 70.54 & 70.98 & 73.65  \\
  CC \cite{peng2019correlation}  & 71.42 & 70.96 & 73.76 \\
  SP \cite{tung2019similarity} & 71.59 & 71.15 & 73.95 \\
 PKT \cite{pkt}  & 71.68 & 71.08 & 74.01  \\
$r^f+\mathcal{L}_2$ &  71.93 & 71.54 & 74.15 \\
$r^f+\mathcal{L}_{RC}$ &  72.70 &72.02 & 74.69 \\
\bottomrule
\end{tabular}
    \label{tab:relationvalidate}
    \vspace{-0.2em}
\end{table}

\begin{figure}
    \centering
    \includegraphics[width=0.45\textwidth]{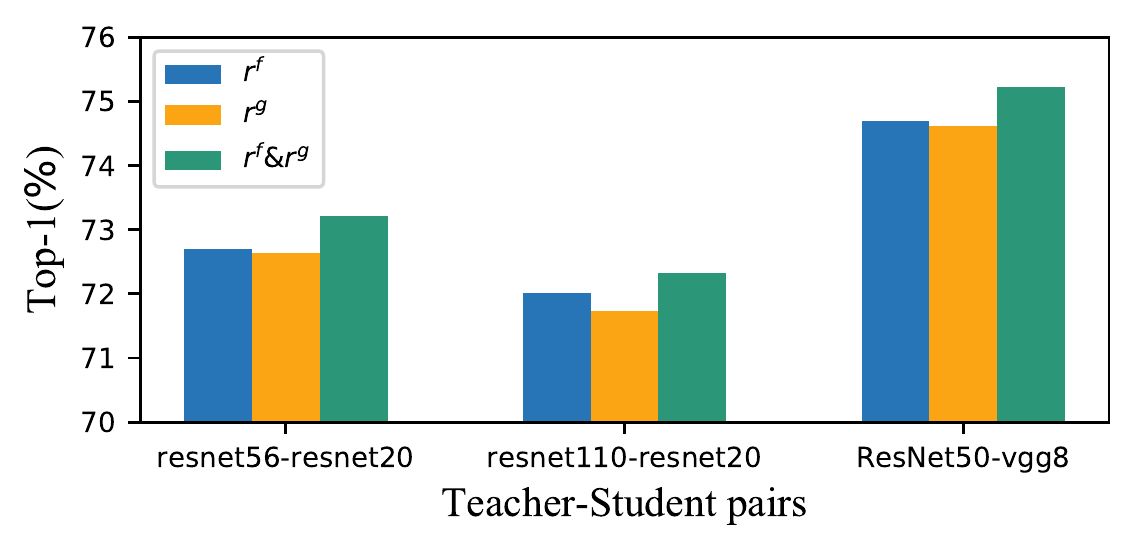}
    \caption{Accuracy of different relation elements.  The feature relation $r^f$, gradident relation $r^g$, and complementary relation $r^f\&r^g$  are distilled on three teacher-student pairs.}
    \label{fig:relation_complementary}
\end{figure}

\noindent \textbf{Parameter Setting.} 
For CIFAR, mini-batch size is set to 64 in 1 GPU.
 SGD optimizer is used  with weight decay and momentum of 0.0001 and 0.9 respectively.
And the learning rate  and schedule strategy  follow \cite{crd}, which is included in supplementary materials.
For ImageNet, batchsize is set to 256 in 8 GPUs, and the standard training settings for ImageNet is adopted.
For other competing methods, we use the implementation settings in papers or official shared codes.
The relation dimension computed by sub-networks $M^{T,S}$ and $M^{T}$ is set to 256-d since the representation dimension in most of our experimental networks is 256-d.

\subsection{Ablation Study}
\label{ablation}
 Three  teacher-student pairs are selected  for ablation study. Their model names and 
top-1 accuracy (\%) when trained individually on CIFAR100 are shown below:
\begin{table}[h]
    \centering
    \small
\begin{tabular}{c|cc|c}
\toprule
\multirow{2}{*}{teacher}   & resnet56 &resnet110 &  ResNet50 \\
   & 73.25 & 73.89 & 79.04 \\ 
  \midrule 
\multirow{2}{*}{student}  & resnet20 & resnet20 & vgg8\\
   & 69.06 & 69.06 & 70.71 \\
\bottomrule
\end{tabular}

\end{table}

\noindent The first two are with similar architectures, while the last one is with a  very different architecture. 
 These experiments are conducted on CIFAR100, and results are averaged over 3 runs.

\noindent \textbf{Effectiveness of relation modeling method.}
\label{exp_relation}
We first demonstrate the effectiveness of anchor-based relation. In contrast to conventional modeling methods, our relation is cross-space and high-dimensional. To verify its superiority, we compare it with four methods using low-dimensional relations:
1) RKD \cite{park2019relational}; 2) CC \cite{peng2019correlation}; 3) SP \cite{tung2019similarity}; and 4) PKT \cite{pkt}.
For a fair comparison, we also use $\mathcal{L}_2$ loss to preserve representation relations and only feature relation is involved.
The results are shown in Tab. \ref{tab:relationvalidate}. 
Over all three teacher-student pairs, our proposed relation  boosts the test accuracy by a large margin even with $\mathcal{L}_2$ loss, which means that our relation modelling method is superior. 

\noindent \textbf{Effectiveness of complementary relation elements.}
We propose two elements: feature  and its gradient, to model representation relation.
To verify  their  complementarity, we test the  distilling  accuracy of these two elements when used alone and when used simultaneously.
As Fig. \ref{fig:relation_complementary} shows, their combination can get the best result, which indicates that the feature and the gradient
are complementary to each other and can more comprehensively present the representation interdependences.

\begin{table}[t]
\small
    \centering
\caption{Testing accuracy (\%) on CIFAR100 with different transformations for critic function $h$. $IM$: identity mapping; $LP$: linear projection; $NP$: nonlinear  projection. 
The transformation dimensions are appended as subscripts.}
\begin{tabular}{c|cc|c}
\toprule
teacher & resnet56 &resnet110 &  ResNet50 \\
student   &  resnet20 & resnet20 & vgg8\\
\midrule
$IM$ & 72.35  & 71.84 & 74.25 \\

$NP_{256}$ & 72.52 & 71.98 & 74.49 \\
\midrule
$LP_{64}$ &72.45 & 71.92 & 74.34 \\
$LP_{128}$ & \textbf{72.70} & 72.02 & \textbf{74.69}\\
$LP_{256}$ & 72.65 & \textbf{72.12} & 74.57 \\
\bottomrule
\end{tabular}
    
    \label{tab:critic}
\end{table}
\begin{figure}[t]
     \centering
     \begin{subfigure}[b]{0.23\textwidth}
         \centering
         \includegraphics[width=\textwidth]{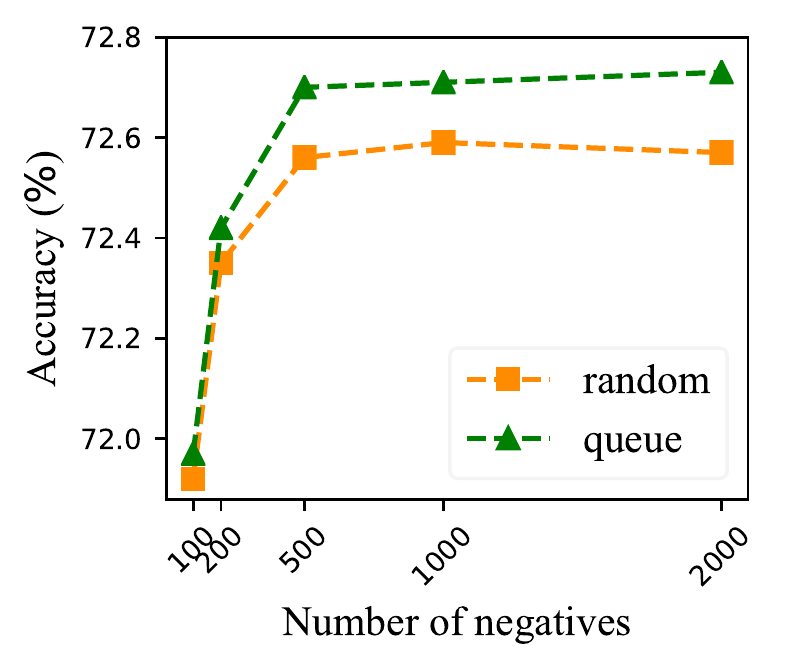}
         \caption{Effects of varying $N$}
         \label{fig: negative number}
     \end{subfigure}
     \hfill
      \hspace{.000001in}
     \begin{subfigure}[b]{0.23\textwidth}
         \centering
         \includegraphics[width=\textwidth]{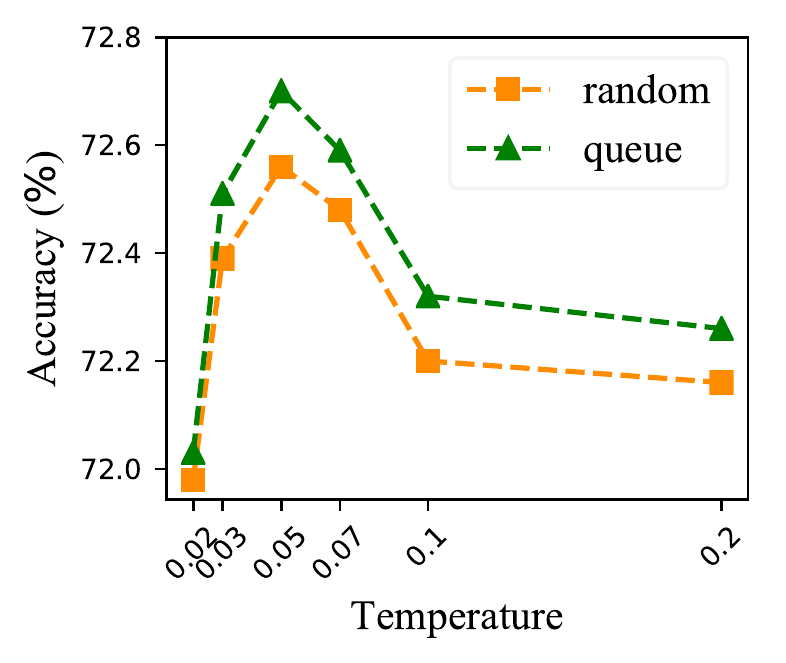}
         \caption{Effects of varying $\tau$}
         \label{fig:temperature}
     \end{subfigure}
        \caption{Accuracy of varying negative number $N$ and temperature $\tau$ with different sample policies.}
        \label{fig:queue}
\end{figure}

\noindent \textbf{Effectiveness of critic function $h$.}
We propose the critic function $h$ in Eq. \ref{hfunction} to estimate the distribution $q(C=1|R^T,R^{T,S})$.
 To investigate the effectiveness of $h_{1}$ and $h_{2}$ selection, we conduct three experiments, including specifying the $h_1$ and $h_2$ functions with identity mapping, nonlinear projection and linear transformation(default).
 In particular, $h$ is degraded to cosine similarity estimation when identity mapping is adopted. For nonlinear projection, we use a MLP with one hidden layer $h(r)=W^{(2)}\sigma(W^{(1)}r)$ where $r$ is input relation and $\sigma$ is a ReLU nonlinearity.

In this study, the output dimension of linear or nonlinear transformation are both 256.
Table \ref{tab:critic} shows testing results using different transformations. We observe that both the linear and nonlinear projection achieve better results than identity mapping under the same projection dimension, which means that critic function with learnable parameters can better fit the distribution $q(C=1|R^T,R^{T,S})$.

\begin{table}[t]
    \centering
    \small
    \caption{Contrastive loss functions. To simplify, the
 anchor relation $r^{T}_{ij}$, positive relation $r^{T,S}_{ij}$, and negative relation $r^{T,S}_{ik_{j\neq k}}$  after critic transformation are denoted as $u$, $v^{+}$ and $v^{-}$ respectively. All relations are $\ell_2$ normalized  before inner product. $\tau$ is the temperature weight, and $m$ is the margin parameter. Additionally, $\sigma$ is \textit{sigmoid} function.  }
    \begin{tabular}{c|c}
    \toprule
       Name  &  Loss function \\
    \midrule
        $\mathcal{L}_{MT} \cite{marginLOss}$ & $max(u^{'}v^{-}-u^{'}v^{+}+m,0)$ \\[1ex]
        $\mathcal{L}_{CL}$ \cite{li2014deepreid}& $- \log{\sigma(u^{'}v^{+}/\tau)} - \log{(1-\sigma(u^{'}v^{-}/\tau))}$ \\[1ex]
        $\mathcal{L}_{NCE}$ \cite{infoNCE} &  $ -  u^{'}v^{+}/\tau + \log{\sum e^{ u^{'}v^{-}/\tau}} $ \\[1ex]
        $\mathcal{L}_{RC}$ & $-\log{{\frac{e^{u^{'}v^{+}/\tau}}{e^{1/\tau}}} } - N \sum \log{(1 - {\frac{e^{u^{'}v^{-}/\tau}}{e^{1/\tau}}} )} $\\[0.5ex]
        
    \bottomrule
    
    \end{tabular}
    \label{tab:contrasiveloss}
\end{table}
\begin{table}[t]
\small
    \centering
    \caption{Testing accuracy (\%) on CIFAR100 with different contrastive loss functions.}
\begin{tabular}{c c|cc|c}
\toprule
teacher & & resnet56 &resnet110 &  ResNet50 \\
student &  &  resnet20 & resnet20 & vgg8\\
\midrule
$\mathcal{L}_2$ & & 71.93 & 71.54 & 73.89 \\
 
$\mathcal{L}_{MT}$ & $m=0.4$ & 72.21 & 71.83 & 74.25   \\
$\mathcal{L}_{CL}$ &$\tau=0.05$ & 72.15 & 71.72 & 74.07 \\
$\mathcal{L}_{NCE}$& $\tau=0.05$ & 72.53 & \textbf{72.09} & 74.44 \\
$\mathcal{L}_{RC}$& $\tau=0.05$ & \textbf{72.70 }& 72.02 & \textbf{74.69} \\
\bottomrule
\end{tabular}
    \label{tab:differentloss}
\end{table}
\noindent \textbf{Effectiveness of relation contrastive loss.}
We compare our relation contrastive loss $\mathcal{L}_{RC}$ with other commonly used contrastive loss, such as triplet loss with margin ($\mathcal{L}_{MT}$) \cite{marginLOss} and contrastive logistic loss ($\mathcal{L}_{CL}$) \cite{li2014deepreid, pair_loss}.
Tab. \ref{tab:contrasiveloss} shows the formulations of four contrastive loss function.

To better analyze the loss function, we only use the feature element and gradient is not employed. The hyperparameters in these losses, \emph{i.e.,}  temperature $\tau$ and margin $m$, are tuned to achieve the best results.
Results reported in Tab. \ref{tab:differentloss} show that, $\mathcal{L}_{NCE}$ and $\mathcal{L}_{RC}$ can significantly outperform  $\mathcal{L}_{MT}$ and $\mathcal{L}_{CL}$, because they can benefit from large number of negative samples. While our objective function $\mathcal{L}_{RC}$ is better than $\mathcal{L}_{NCE}$ in most of teacher-student combinations.

    

    

\subsection{Hyper-parameter analysis}

\begin{table*}[th]
\small
    \centering
    \caption{ \textbf{The top-1 accuracies ($\%$) of seven different  student-teacher pairs on CIFAR100.}
The accuracies of the teachers' and students' performance when they are trained individually are presented in the second partition after the header.
FRCD (or GRCD) is the incomplete version of CRCD which  means that only feature relation (or gradient relation) is employed in distillation. The best results are \textbf{bolded} and the best in competing methods are \underline{underlined}.}
    \begin{tabular}{lcccccccccc}
\toprule
Teacher & WRN-40-2 & WRN-40-2 & resnet56 & resnet110 & resnet110 & resnet32x4 & vgg13  \\
Student & WRN-16-2 & WRN-40-1 & resnet20 & resnet20 & resnet32 & resnet8x4 & vgg8 \\
\midrule
Teacher & 76.64 & 76.64 & 73.25 & 73.89 & 73.89 & 79.61 & 75.00 \\
Student & 73.53 & 72.33 & 69.06 & 69.06 & 72.31 & 72.57 & 70.71 \\
\midrule
KD \cite{hinton2015distilling}    & 75.11 & 73.59 & 71.08 & 70.92 & 73.07 & 73.19 & 72.85 \\
FitNet \cite{romero2014fitnets} & 75.37 & 73.71 & 71.65 & 70.95 & 73.21 & 73.42 & 73.24 \\
AT \cite{ATloss}    & 75.92 & 73.92 & 71.69 & 71.03 & 73.29 & 73.29 & 73.16 \\
SP \cite{tung2019similarity}    & 75.84 & 73.85 & 71.59 & 71.15 & 73.12 & 73.36 & 73.29 \\
CC  \cite{peng2019correlation}   & 75.89 & 73.69 & 71.42 & 70.96 & 73.06 & 73.52 & 73.06 \\
VID  \cite{vid}  & 75.53 & 73.95 & 71.32 & 70.93 & 73.19 & 73.75 & 73.13 \\
RKD \cite{park2019relational}  & 75.20 & 73.76 & 71.54 & 70.98 & 73.25 & 73.51 & 73.09 \\
PKT \cite{pkt}    & 75.67 & 73.89 & 71.68 & 71.08 & 73.32 & 73.63 & 73.28 \\
AB  \cite{ab}   & 71.31 & 73.76 & 71.29 & 70.95 & 73.16 & 73.43 & 73.02 \\
FT \cite{factor}   & 75.78 & 74.02 & 71.52 & 71.03 & 73.21 & 73.28 & 73.19 \\
NST \cite{NST}   & 74.51 & 73.62 & 71.47 & 71.14 & 73.21 & 73.58 & 73.14 \\
CRD  \cite{crd}  & \underline{75.97} & 74.47 & \underline{71.75} & \underline{71.52} & \underline{73.81} & \underline{75.62} & 74.42 \\
SSKD \cite{SSKD}  & 75.39 & \underline{75.30} & 70.29 & 71.48 & 73.64 & 75.53 & \underline{74.51} \\
\midrule
FRCD & 76.18 & 75.26 & 72.70 & 72.02 & 74.65 & 75.99 & 74.54 \\
GRCD & 76.27 & 75.24 & 72.64 & 71.73 & 74.48 & 75.57 & 74.32 \\
CRCD   &  \textbf{ 76.67} &\textbf{ 75.95} &\textbf{ 73.21} & \textbf{72.33} & \textbf{74.98} & \textbf{76.42} & \textbf{74.97}\\
\bottomrule
\end{tabular}
    \vspace{-0.2em}
    \label{tab:resultwithsot}
\end{table*}

\begin{table*}[t]
\small
    \centering
    \caption{\textbf{Top-1 and Top-5 error rate ($\%$)  on ImageNet validation set.} We compare our CRCD with competing methods including AT \cite{ATloss}, KD\cite{hinton2015distilling}, SP \cite{tung2019similarity}, CC \cite{peng2019correlation}, CRD \cite{crd} and  SSKD \cite{SSKD}, and folow the training settings in \cite{crd}. 
    }
    \begin{tabular}{c|cc|cccccc|c}
\toprule
 & Teacher & Student & AT & KD & SP & CC  & CRD & SSKD & CRCD \\
 \midrule
 Top-1 & 26.69 & 30.25 & 29.30 & 29.34 & 29.38 & 30.04 & 28.62 & 28.38 & $\mathbf{ 28.04}$ \\
Top-5 & 8.58 & 10.93 & 10.00 & 10.12 & 10.20 & 10.83 & 9.51 & 9.33 & $\mathbf{9.06}$ \\
\bottomrule
\end{tabular}
    \vspace{-0.4em}
    \label{tab:imagenet}
\end{table*}

Several hyper-parameters are worth investigating in our proposed CRCD method. (1) The number of negative samples $N$; 
(2) The temperature used to scale the critic scores in Eq. \ref{hfunction}; (3) The sampling policy to construct negative relations; (4) The projection dimension of critic function $h$.
We adopt resnet56-resnet20 pair on CIFAR100 for analysis.

\noindent \textbf{Number of negative samples.} We validate different $N$: 100,  200,  500, 1000, 2000. As shown in Tab. \ref{fig: negative number}, increasing the negative number leads to better performance, and the performance is saturated when $n > 500$. We therefore utilize $N=500$ in all other experiments to save computational cost. Compared to  CRD \cite{crd}, our CRCD requires fewer negative features to reduce the need of memory. This is because CRCD can utilize few samples to generate a large number of relations, while CRD only depends on the number of samples. 


\noindent \textbf{Temperature $\tau$.} Fig. \ref{fig:temperature} 
reports the results when $\tau$ varies from 0.02 to 0.2. We find that both extremely high or low temperature leads to inferior performance.
In general, a temperature  between 0.03 to 0.07  works well. We set $\tau=0.05$ for all other 
experiments.

\noindent \textbf{Sampling policy.}
To ensure that negative samples are as up-to-date as possible,  we store features and gradients in a queue way which will remove the oldest sample when adding the latest sample. 
We compare the randomly sampling policy and the queuing sampling policy in Fig. \ref{fig:queue}. The queuing sampling policy (denoted as queue) can consistently outperform the naive randomly sampling policy (denoted as random) when varying negative number $N$ and temperature $\tau$.

\noindent \textbf{Projection dimension.} 
We investigate the influence of output dimension for critic function  $h$ by setting output dimension to 64, 128, and 256 (the input relation dimension is  256-d).
As shown in Tab. \ref{tab:critic},
 compared to 128-d or 256-d,  transformation with  lower dimension (64-d)  has some accuracy degradation. We utilize the 128-d linear transformation to make a trade-off between effectiveness and computational cost.

\subsection{Comparison with State-of-the-arts}

\noindent \textbf{CIFAR100.} 
We compare our CRCD with other advanced knowledge distillation methods in Tab. \ref{tab:resultwithsot}.
 Various modern CNN architectures \cite{resnet,mobilenets,shufflenet,wideresnet}
are selected as teacher networks or student networks.
For a fair comparison, we combine all distillation methods with conventional KD \cite{hinton2015distilling}. From Tab. \ref{tab:resultwithsot}, we can observe that our distillation method CRCD can consistently outperform all other distillation methods with a large margin, including the recent state-of-the-arts, CRD and SSKD.
Additionally, even only one element (feature or its gradient)
is used in the relation distillation, our method can still achieve the competing accuracy when compared to CRD or SSKD. When the feature and its gradient are employed in the representation relation distillation simultaneously, our CRCD can significantly outperform the other methods. In particular, the accuracy gap between CRCD and the other best performing method is $0.9\%$ (averaged over 7 
pairs in Tab. \ref{tab:resultwithsot}).

To evaluate the distillation effectiveness across very different network architectures, we also
carry out detailed comparisons in supplementary materials.

\noindent \textbf{ImageNet.}
Following \cite{crd,SSKD}, we adopt the ResNet34-ResNet18 pair to evaluate the effectiveness of CRCD on ImageNet.
As shown in Tab. \ref{tab:imagenet},  the Top-1 and Top-5 accuracy between the teacher and student without  distillation  is $3.56\%$ and $2.43\%$. Our CRCD reaches the best distillation  performance by narrowing the performance gap by $2.21\%$ and $1.87\%$ respectively.
Results on ImageNet demonstrates the scalability of our CRCD to large-scale benchmarks.

\section{Conclusion}
\vspace{-0.6em}
In this work, we have proposed a novel knowledge distillation method, CRCD,  to distill important structural information from a teacher to a student. To better distill the relation knowledge, two sub-networks are used to estimate the cross-space relation and teacher-space relation, respectively. We maximized the mutual information between the two kinds of relations by a newly proposed relation contrastive distillation loss, and utilized two complementary elements, the feature and its gradient, to enhance the representative ability of the relation. With the design of the loss function, the inter-sample relation and representation learning can be optimized simultaneously. Extensive experiments demonstrate the effectiveness of our approach and suggest that the  structural information of deep representation can be better exploited during distillation.

{\small
\bibliographystyle{ieee_fullname}
\bibliography{egbib}

\begin{thebibliography}{10}\itemsep=-1pt

\bibitem{vid}
Sungsoo Ahn, Shell~Xu Hu, Andreas Damianou, Neil~D Lawrence, and Zhenwen Dai.
\newblock Variational information distillation for knowledge transfer.
\newblock In {\em Proceedings of the IEEE Conference on Computer Vision and
  Pattern Recognition}, pages 9163--9171, 2019.

\bibitem{belagiannis2018adversarial}
Vasileios Belagiannis, Azade Farshad, and Fabio Galasso.
\newblock Adversarial network compression.
\newblock In {\em Proceedings of the European Conference on Computer Vision
  (ECCV)}, pages 0--0, 2018.

\bibitem{mine}
Mohamed~Ishmael Belghazi, Aristide Baratin, Sai Rajeswar, Sherjil Ozair, Yoshua
  Bengio, Aaron Courville, and R~Devon Hjelm.
\newblock Mine: mutual information neural estimation.
\newblock {\em arXiv preprint arXiv:1801.04062}, 2018.

\bibitem{caron2020unsupervised}
Mathilde Caron, Ishan Misra, Julien Mairal, Priya Goyal, Piotr Bojanowski, and
  Armand Joulin.
\newblock Unsupervised learning of visual features by contrasting cluster
  assignments.
\newblock {\em arXiv preprint arXiv:2006.09882}, 2020.

\bibitem{Chen_2018_CVPR_video}
Dapeng Chen, Hongsheng Li, Tong Xiao, Shuai Yi, and Xiaogang Wang.
\newblock Video person re-identification with competitive snippet-similarity
  aggregation and co-attentive snippet embedding.
\newblock In {\em Proceedings of the IEEE Conference on Computer Vision and
  Pattern Recognition (CVPR)}, June 2018.

\bibitem{Chen_2018_CVPR}
Dapeng Chen, Dan Xu, Hongsheng Li, Nicu Sebe, and Xiaogang Wang.
\newblock Group consistent similarity learning via deep crf for person
  re-identification.
\newblock In {\em Proceedings of the IEEE Conference on Computer Vision and
  Pattern Recognition (CVPR)}, June 2018.

\bibitem{Chen_2016_CVPR}
Dapeng Chen, Zejian Yuan, Badong Chen, and Nanning Zheng.
\newblock Similarity learning with spatial constraints for person
  re-identification.
\newblock In {\em Proceedings of the IEEE Conference on Computer Vision and
  Pattern Recognition (CVPR)}, June 2016.

\bibitem{SimCLR}
Ting Chen, Simon Kornblith, Mohammad Norouzi, and Geoffrey Hinton.
\newblock A simple framework for contrastive learning of visual
  representations.
\newblock {\em arXiv preprint arXiv:2002.05709}, 2020.

\bibitem{chen2020big}
Ting Chen, Simon Kornblith, Kevin Swersky, Mohammad Norouzi, and Geoffrey
  Hinton.
\newblock Big self-supervised models are strong semi-supervised learners.
\newblock {\em arXiv preprint arXiv:2006.10029}, 2020.

\bibitem{imagenet}
Jia Deng, Wei Dong, Richard Socher, Li-Jia Li, Kai Li, and Li Fei-Fei.
\newblock Imagenet: A large-scale hierarchical image database.
\newblock In {\em 2009 IEEE conference on computer vision and pattern
  recognition}, pages 248--255. Ieee, 2009.

\bibitem{pair_loss}
Weijian Deng, Liang Zheng, Qixiang Ye, Guoliang Kang, Yi Yang, and Jianbin
  Jiao.
\newblock Image-image domain adaptation with preserved self-similarity and
  domain-dissimilarity for person re-identification.
\newblock In {\em Proceedings of the IEEE Conference on Computer Vision and
  Pattern Recognition (CVPR)}, June 2018.

\bibitem{ge2020mutual}
Yixiao Ge, Dapeng Chen, and Hongsheng Li.
\newblock Mutual mean-teaching: Pseudo label refinery for unsupervised domain
  adaptation on person re-identification.
\newblock In {\em International Conference on Learning Representations}, 2020.

\bibitem{ge2020selfpaced}
Yixiao Ge, Feng Zhu, Dapeng Chen, Rui Zhao, and Hongsheng Li.
\newblock Self-paced contrastive learning with hybrid memory for domain
  adaptive object re-id.
\newblock In {\em Advances in Neural Information Processing Systems}, 2020.

\bibitem{gou2020knowledge}
Jianping Gou, Baosheng Yu, Stephen~John Maybank, and Dacheng Tao.
\newblock Knowledge distillation: A survey.
\newblock {\em arXiv preprint arXiv:2006.05525}, 2020.

\bibitem{byol}
Jean-Bastien Grill, Florian Strub, Florent Altch{\'e}, Corentin Tallec,
  Pierre~H Richemond, Elena Buchatskaya, Carl Doersch, Bernardo~Avila Pires,
  Zhaohan~Daniel Guo, Mohammad~Gheshlaghi Azar, et~al.
\newblock Bootstrap your own latent: A new approach to self-supervised
  learning.
\newblock {\em arXiv preprint arXiv:2006.07733}, 2020.

\bibitem{moco}
Kaiming He, Haoqi Fan, Yuxin Wu, Saining Xie, and Ross Girshick.
\newblock Momentum contrast for unsupervised visual representation learning.
\newblock In {\em Proceedings of the IEEE/CVF Conference on Computer Vision and
  Pattern Recognition}, pages 9729--9738, 2020.

\bibitem{resnet}
Kaiming He, Xiangyu Zhang, Shaoqing Ren, and Jian Sun.
\newblock Deep residual learning for image recognition.
\newblock In {\em Proceedings of the IEEE conference on computer vision and
  pattern recognition}, pages 770--778, 2016.

\bibitem{ab}
Byeongho Heo, Minsik Lee, Sangdoo Yun, and Jin~Young Choi.
\newblock Knowledge transfer via distillation of activation boundaries formed
  by hidden neurons.
\newblock In {\em Proceedings of the AAAI Conference on Artificial
  Intelligence}, volume~33, pages 3779--3787, 2019.

\bibitem{hinton2015distilling}
Geoffrey Hinton, Oriol Vinyals, and Jeff Dean.
\newblock Distilling the knowledge in a neural network.
\newblock {\em arXiv preprint arXiv:1503.02531}, 2015.

\bibitem{DeepInfoMAx}
R~Devon Hjelm, Alex Fedorov, Samuel Lavoie-Marchildon, Karan Grewal, Phil
  Bachman, Adam Trischler, and Yoshua Bengio.
\newblock Learning deep representations by mutual information estimation and
  maximization.
\newblock {\em arXiv preprint arXiv:1808.06670}, 2018.

\bibitem{mobilenets}
Andrew~G Howard, Menglong Zhu, Bo Chen, Dmitry Kalenichenko, Weijun Wang,
  Tobias Weyand, Marco Andreetto, and Hartwig Adam.
\newblock Mobilenets: Efficient convolutional neural networks for mobile vision
  applications.
\newblock {\em arXiv preprint arXiv:1704.04861}, 2017.

\bibitem{NST}
Zehao Huang and Naiyan Wang.
\newblock Like what you like: Knowledge distill via neuron selectivity
  transfer.
\newblock {\em arXiv preprint arXiv:1707.01219}, 2017.

\bibitem{factor}
Jangho Kim, SeongUk Park, and Nojun Kwak.
\newblock Paraphrasing complex network: Network compression via factor
  transfer.
\newblock In {\em Advances in neural information processing systems}, pages
  2760--2769, 2018.

\bibitem{cifar100}
Alex Krizhevsky, Geoffrey Hinton, et~al.
\newblock Learning multiple layers of features from tiny images.
\newblock 2009.

\bibitem{Li_2019_ICCV}
Suichan Li, Dapeng Chen, Bin Liu, Nenghai Yu, and Rui Zhao.
\newblock Memory-based neighbourhood embedding for visual recognition.
\newblock In {\em Proceedings of the IEEE/CVF International Conference on
  Computer Vision (ICCV)}, October 2019.

\bibitem{li2014deepreid}
Wei Li, Rui Zhao, Tong Xiao, and Xiaogang Wang.
\newblock Deepreid: Deep filter pairing neural network for person
  re-identification.
\newblock In {\em Proceedings of the IEEE conference on computer vision and
  pattern recognition}, pages 152--159, 2014.

\bibitem{liu2019knowledge}
Iou-Jen Liu, Jian Peng, and Alexander~G Schwing.
\newblock Knowledge flow: Improve upon your teachers.
\newblock {\em arXiv preprint arXiv:1904.05878}, 2019.

\bibitem{relationgraph}
Yufan Liu, Jiajiong Cao, Bing Li, Chunfeng Yuan, Weiming Hu, Yangxi Li, and
  Yunqiang Duan.
\newblock Knowledge distillation via instance relationship graph.
\newblock In {\em Proceedings of the IEEE Conference on Computer Vision and
  Pattern Recognition}, pages 7096--7104, 2019.

\bibitem{malinin2019ensemble}
Andrey Malinin, Bruno Mlodozeniec, and Mark Gales.
\newblock Ensemble distribution distillation.
\newblock {\em arXiv preprint arXiv:1905.00076}, 2019.

\bibitem{CPC}
Aaron van~den Oord, Yazhe Li, and Oriol Vinyals.
\newblock Representation learning with contrastive predictive coding.
\newblock {\em arXiv preprint arXiv:1807.03748}, 2018.

\bibitem{infoNCE}
Aaron van~den Oord, Yazhe Li, and Oriol Vinyals.
\newblock Representation learning with contrastive predictive coding.
\newblock {\em arXiv preprint arXiv:1807.03748}, 2018.

\bibitem{park2019relational}
Wonpyo Park, Dongju Kim, Yan Lu, and Minsu Cho.
\newblock Relational knowledge distillation.
\newblock In {\em Proceedings of the IEEE Conference on Computer Vision and
  Pattern Recognition}, pages 3967--3976, 2019.

\bibitem{pkt}
Nikolaos Passalis and Anastasios Tefas.
\newblock Learning deep representations with probabilistic knowledge transfer.
\newblock In {\em Proceedings of the European Conference on Computer Vision
  (ECCV)}, pages 268--284, 2018.

\bibitem{peng2019correlation}
Baoyun Peng, Xiao Jin, Jiaheng Liu, Dongsheng Li, Yichao Wu, Yu Liu, Shunfeng
  Zhou, and Zhaoning Zhang.
\newblock Correlation congruence for knowledge distillation.
\newblock In {\em Proceedings of the IEEE International Conference on Computer
  Vision}, pages 5007--5016, 2019.

\bibitem{romero2014fitnets}
Adriana Romero, Nicolas Ballas, Samira~Ebrahimi Kahou, Antoine Chassang, Carlo
  Gatta, and Yoshua Bengio.
\newblock Fitnets: Hints for thin deep nets.
\newblock {\em arXiv preprint arXiv:1412.6550}, 2014.

\bibitem{marginLOss}
Florian Schroff, Dmitry Kalenichenko, and James Philbin.
\newblock Facenet: A unified embedding for face recognition and clustering.
\newblock In {\em Proceedings of the IEEE Conference on Computer Vision and
  Pattern Recognition (CVPR)}, June 2015.

\bibitem{shen2019meal}
Zhiqiang Shen, Zhankui He, and Xiangyang Xue.
\newblock Meal: Multi-model ensemble via adversarial learning.
\newblock In {\em Proceedings of the AAAI Conference on Artificial
  Intelligence}, volume~33, pages 4886--4893, 2019.

\bibitem{MutualCRF}
Shixiang Tang, Dapeng Chen, Lei Bai, Yixiao Ge, and Wanli Ouyang.
\newblock Mutual crf-gnn for few shot learning.
\newblock In {\em Proceedings of the IEEE Conference on Computer Vision and
  Pattern Recognition}, 2021.

\bibitem{LayerwiseCTL}
Shixiang Tang, Dapeng Chen, Jinguo Zhu, Shijie Yu, and Wanli Ouyang.
\newblock Layerwise optimization by gradient decomposition for continual
  learning.
\newblock In {\em Proceedings of the IEEE Conference on Computer Vision and
  Pattern Recognition}, 2021.

\bibitem{Contrastive_multiview_coding}
Yonglong Tian, Dilip Krishnan, and Phillip Isola.
\newblock Contrastive multiview coding.
\newblock {\em arXiv preprint arXiv:1906.05849}, 2019.

\bibitem{crd}
Yonglong Tian, Dilip Krishnan, and Phillip Isola.
\newblock Contrastive representation distillation.
\newblock {\em arXiv preprint arXiv:1910.10699}, 2019.

\bibitem{tian2020makes}
Yonglong Tian, Chen Sun, Ben Poole, Dilip Krishnan, Cordelia Schmid, and
  Phillip Isola.
\newblock What makes for good views for contrastive learning.
\newblock {\em arXiv preprint arXiv:2005.10243}, 2020.

\bibitem{tung2019similarity}
Frederick Tung and Greg Mori.
\newblock Similarity-preserving knowledge distillation.
\newblock In {\em Proceedings of the IEEE International Conference on Computer
  Vision}, pages 1365--1374, 2019.

\bibitem{wang2020understanding}
Tongzhou Wang and Phillip Isola.
\newblock Understanding contrastive representation learning through alignment
  and uniformity on the hypersphere.
\newblock {\em arXiv preprint arXiv:2005.10242}, 2020.

\bibitem{wu2019distilled}
Ancong Wu, Wei-Shi Zheng, Xiaowei Guo, and Jian-Huang Lai.
\newblock Distilled person re-identification: Towards a more scalable system.
\newblock In {\em Proceedings of the IEEE Conference on Computer Vision and
  Pattern Recognition}, pages 1187--1196, 2019.

\bibitem{wu2018unsupervised}
Zhirong Wu, Yuanjun Xiong, Stella~X Yu, and Dahua Lin.
\newblock Unsupervised feature learning via non-parametric instance
  discrimination.
\newblock In {\em Proceedings of the IEEE Conference on Computer Vision and
  Pattern Recognition}, pages 3733--3742, 2018.

\bibitem{SSKD}
Guodong Xu, Ziwei Liu, Xiaoxiao Li, and Chen~Change Loy.
\newblock Knowledge distillation meets self-supervision.
\newblock {\em arXiv preprint arXiv:2006.07114}, 2020.

\bibitem{xu2017training}
Zheng Xu, Yen-Chang Hsu, and Jiawei Huang.
\newblock Training shallow and thin networks for acceleration via knowledge
  distillation with conditional adversarial networks.
\newblock {\em arXiv preprint arXiv:1709.00513}, 2017.

\bibitem{yim2017gift}
Junho Yim, Donggyu Joo, Jihoon Bae, and Junmo Kim.
\newblock A gift from knowledge distillation: Fast optimization, network
  minimization and transfer learning.
\newblock In {\em Proceedings of the IEEE Conference on Computer Vision and
  Pattern Recognition}, pages 4133--4141, 2017.

\bibitem{fsp}
Junho Yim, Donggyu Joo, Jihoon Bae, and Junmo Kim.
\newblock A gift from knowledge distillation: Fast optimization, network
  minimization and transfer learning.
\newblock In {\em Proceedings of the IEEE Conference on Computer Vision and
  Pattern Recognition}, pages 4133--4141, 2017.

\bibitem{ATloss}
Sergey Zagoruyko and Nikos Komodakis.
\newblock Paying more attention to attention: Improving the performance of
  convolutional neural networks via attention transfer.
\newblock {\em arXiv preprint arXiv:1612.03928}, 2016.

\bibitem{wideresnet}
Sergey Zagoruyko and Nikos Komodakis.
\newblock Wide residual networks.
\newblock {\em arXiv preprint arXiv:1605.07146}, 2016.

\bibitem{shufflenet}
Xiangyu Zhang, Xinyu Zhou, Mengxiao Lin, and Jian Sun.
\newblock Shufflenet: An extremely efficient convolutional neural network for
  mobile devices.
\newblock In {\em Proceedings of the IEEE conference on computer vision and
  pattern recognition}, pages 6848--6856, 2018.

\bibitem{crowded}
Jinguo Zhu, Zejian Yuan, Chong Zhang, Wanchao Chi, Yonggen Ling, et~al.
\newblock Crowded human detection via an anchor-pair network.
\newblock In {\em Proceedings of the IEEE/CVF Winter Conference on Applications
  of Computer Vision}, pages 1391--1399, 2020.

\end{thebibliography}
}

\end{document}